\documentclass{article}
\usepackage{ijcai21}

\usepackage{times}

\usepackage{soul}
\usepackage{url}
\usepackage[hidelinks]{hyperref}
\usepackage[utf8]{inputenc}
\usepackage[small]{caption}
\usepackage{graphicx}
\usepackage{amsmath}
\usepackage{booktabs}
\usepackage{todonotes}
\usepackage[acronym]{glossaries}
\newacronym{aiml}{AIML}{Artificial Intelligence and Machine Learning}
\newacronym{pocus}{POCUS}{Point of Care Ultrasounds}

\title{Next-Gen Machine Learning Supported Diagnostic Systems for Spacecraft}

\author{
Athanasios Vlontzos$^1$\footnote{Contact Author \\ Accepted in the AI for Spacecraft Longevity Workshop at IJCAI 2021 }\and
Gabriel Sutherland$^2$\and
Siddha Ganju $^{3}$\And
Frank Soboczenski$^{4}$
\\
\affiliations
$^1$Imperial College London,
$^2$Oregon State University, 
$^3$NVIDIA Corp,
$^4$King's College London\\
\emails
athanasios.vlontzos14@ic.ac.uk,
sutherlg@oregonstate.edu,
sganju@nvidia.com,
frank.soboczenski@kcl.ac.uk
}

\begin{document}

\maketitle

\begin{abstract}
Future short or long-term space missions require a new generation of monitoring and diagnostic systems due to communication impasses as well as limitations in specialized crew and equipment. Machine learning supported diagnostic systems present a viable solution for medical and technical applications. We discuss challenges and applicability of such systems in light of upcoming missions and outline an example use case for a next-generation medical diagnostic system for future space operations. Additionally, we present approach recommendations and constraints for the successful generation and use of machine learning models aboard a spacecraft.
\end{abstract}

\section{Introduction}
Space agencies have a renewed drive to take human space exploration beyond Low Earth Orbit (LEO) and into deep space. NASA's Artemis program outlines a clear path to return to the Moon and to go beyond to Mars \cite{artemis}. Additionally, recent successes in the commercial space sector by major players such as SpaceX and Blue Origin make human spaceflight more accessible, affordable and future long term-missions a reality. Yet future long duration spaceflight require systems that are independent of LEO operations such as constant communication, the ability to transfer large amounts of data via multiple systems in a relatively short time-frame or the ability to request and exchange crew if needed. On Earth, machine learning (ML) and machine automation is already driving the next industrial revolution and resulted in fully autonomous industrial processes in domains such as agriculture as well as manufacturing \cite{iotagri,iotmanufac}. Spaceflight itself, however, is far behind such advances. Here we discuss challenges ML supported systems face in the space domain as well as the applicability and advantages of ML systems on a spacecraft. We highlight aforementioned items via an example of an autonomous medical system and describe an infrastructure for the successful development of such systems. 




\section{Challenges for ML aboard Spacecraft}
Space is hard and manned space exploration is dangerous and unforgiving. Moreover, the long term effects of human presence in microgravity are still not fully known. Current space missions do rarely include a medical officer among the crew and rely on specialists who are also trained in emergency medicine. While this is sufficient for minor and trained emergency cases, it does not allow for more serious and complex medical treatments. Health related checks and emergencies are handled in a tele-medicine regime where instructions are communicated to the astronauts via ground to spacecraft channels. In extreme emergencies the astronauts can always make the journey back on Earth. As missions veer further away from Earth, returning for medical treatment and relying on simultaneous communications becomes infeasible as both distance and communication latency increases exponentially. For a deployed ML system aboard a spacecraft there are several challenges to consider: a) limited live testing abilities available, and testing in the form of payloads on missions are expensive; b) systems that are deployed need to be at a high technology readiness level (TRL) \cite{alex2021technology}; c) environmental effects may influence deployed systems. For example, how ionizing radiation can affect deployed space capable hardware \cite{1532657}, affecting the  consistency of sensor behavior such as early or late sensor fusion; leading to potentially corrupted data to be processed e) payloads are constrained by weight, so shipping large compute infrastructures is infeasible; f) fully autonomous applications
can be found mainly on controlled environments that assume almost complete access to information and environmental parameters; g) and more importantly the lack of labeled data for each task along with limited interpretability and explainablity of current ML systems add to the complexity. Contrary to that, any space-faring vehicle is faced with extreme environmental conditions that not only are hard to control or predict, but in some cases are challenging to human scientific comprehension. Hence, ML systems must be as robust as possible to changes of their operating environments.

\section{Applicability \& Advantages}
Incorporating autonomous processes in a spacecraft is a complex task as technology and the associated needs constantly develop. Some key points, however, are: (1) Reduce Latency and Earth Reliance - A significant amount of operations aboard modern day spacecrafts and the ISS, require the constant communication with mission control on Earth \cite{dempsey_2017}. As space exploration expands further away from LEO and the Moon, the delay in communication represent an insurmountable obstacle for remote guidance, control and communications. Characteristically, we note that the round-trip time for current communications with Mars ranges between 5 and 20 minutes depending on the state of the two planets orbits \cite{mars2020}. ML, thus, can resolve the dependency on communications and perform the mission critical information processing aboard. (2) Adapting Maintenance - Modern spacecrafts both manned and unmanned, constitute extremely complex systems that even with the use of automated checks are still prone to faults, especially when faced with extraordinary circumstances. The crew might not be able to solve the issues by themselves. To tackle this we believe that ML systems can perform functions that transcend anomaly detection and fault prediction. An ideal deployed ML solution needs to account for automated maintenance and resolution of faults both in hardware and in the software of the spacecraft. 
(3) Reduce latency for functions that don't require manual intervention and can be conducted at scale. For example, automated checksums for data or models (4) Recent advances in ML like bit quantization, pruning, and hardware approximations \cite{Wang_2019} enable inference on resource constraint edge devices which can also be similar to the target hardware in space \cite{9325840}.

\section{Medical Use Case}
In the previous sections we have set out some challenges as well as advantages for the use of ML on spacecrafts. In this section we will be exploring the medical use case. As mentioned before, medical treatment in space relies on communication with the earth, a reliance we need to remove as we progress into deep space missions. A major part of medical treatment is the ability to inspect bodily functions in a non invasive manner through the use of medical imaging devices. Magnetic Resonance (MRI), CT-Scans and X-Rays are widely used on earth but are not ideal for space use, as we explain below. To this effect, we propose the use of \gls{pocus} as a lightweight non ionizing imaging solution.



\noindent\textbf{ML Enabled Imaging:}
Medical imaging devices are often slow, requiring significant resources to operate and store and depend on the use of ionizing radiation (e.g. X-Rays/CT Scans) that has negative effects on patients and doctors alike \cite{fda_xray}. One notable exception to the above constraints is \acrfull{pocus}. Mobile device enabled Ultrasound (US) probes are safe for patients and doctors, require minimal resources to operate ---weighting less than 1kg--- and offer real time imaging capabilities~\cite{gepocus}. However, acquiring medically significant images with a \gls{pocus} probe is non trivial and requires expert operators. 

As expert users might not always be available in deep space missions we envision the use of ML enabled \gls{pocus} probes that guide the user to medically relevant areas. 

Approaching the task of navigation one can identify straight away the need to be able to determine where the intelligent agent is with respect to the patients body. Basic anatomy knowledge by both patient and other crew members is assumed for the purposes of this application. Navigating to points of medical interest, though is non trivial. Following medical practice of first identifying standard planes and anatomical landmarks; reinforcement learning (RL) solutions of disembodied agents have been proposed in \cite{alansary2019evaluating,vlontzos2019multiple} to perform the above tasks, providing high accuracy and low computational constraints. \cite{milletari2019straight,hase2020ultrasoundguided,li2021autonomous} expand the RL methods including the degrees of freedom of the  US probe into the agents action space and optimize directly on identifying landmarks while controlling a virtual probe. All the above methods, find themselves limited to the anatomies that they have been trained on, with their computational burden rising exponentially when trained to find standard planes or landmarks of multiple anatomies.

As such we believe that a hierarchical approach to the problem would help keep complexity and computation low enough for space craft applications while maintaining high enough performance. Approaches like MAX-Q~\cite{dietterich2000hierarchical} have withstood the test of time and have theoretical guarantees on convergence and the ability of the algorithm that given a set of subtasks it can find the global optimal policy. In short, MAX-Q constructs a hierarchical action tree that a higher level agent sets out subtasks for other agents. In this fashion one can design a fully autonomous agent that collects information in form of images through \gls{pocus} navigation, assesses biomarkers based on measurements derived from landmarks, and regresses the diagnosis. 

Furthermore,  \cite{Eslami1204,Dinesh2016,sidekick_RL} have shown that incorporating modules that aid agents "imagine" how scenes would look like from different point of views, increases their performance in scene understanding. Hence, we are firm believers that incorporating such modules in approaches like \cite{milletari2019straight} would increase their performance capabilities. 

Finally, as medical applications are of critical importance, appropriate checks and balances should be put in place to avoid any harm to the crew. We believe that a rule based set of parameters should be developed in collaboration with physicians that would constitute a fallback system. Having no dependence on learned or inferred information we are able to guarantee a basic level of care to the astronauts in case all other systems fail. On top of the safety-net rule based system, a causal inference infrastructure can be used to assess the the probabilities of causality (Sufficiency and Necessity) \cite{buesing2018woulda,kusner2017counterfactual,louizos2017causal,vlontzos2020causal,budd2021}. Causal counterfactual inference enable us to assess the causal links and potential outcomes of treatments providing, thus, a more informed decision process. 

\section{Infrastructure considerations for ML systems on Spacecraft}
Developing novel ML methods to provide mission critical treatments to astronauts is a hard task. In the previous section we set out some thoughts on a potential application for medical imaging aboard a spacecraft. In this section we will be briefly exploring infrastructure considerations that are directly related to the operation and development of medical imaging ML solutions. These considerations are also apparent on earth bound ML applications but gain increased importance due to the edge cases that they are called to operate on during spaceflight.





\noindent\textbf{Data Collection}
Medical data acquisition is faced with a series of challenges. The privacy of the patients donating their data has to be protected throughout the data acquisition and model development processes. In order to comply with all legal obligations, we propose the data to have NIHR~\cite{nihr} approved methods in place for making patient data anonymous. While, storage of data must be in a HIPAA~\cite{hipaa} compliant platform. Federated learning also present promising methods in regards to secure data processing \cite{kon2016}. Going beyond the need of anonymization and maintenance we would like to draw the readers attention to two constraints of increased importance on medical ML applications

\noindent\textbf{\textit{Domain Shift:}} Domain shift robustness is an open topic of ML research that focuses on making existing methods robust to distributional shifts from the underlying training data domain~\cite{long2016unsupervised,zhang2020collaborative}. In medical applications and more prominently in medical imaging applications domain shifts are easier to manifest and harder to overcome. The first major factor in this phenomenon comes from the equipment used. Medical equipment, when installed in hospitals and health centers is configured by the seller to the exact specifications of the attending physicians, as such, two different doctors using the same base equipment on the same patient can result in two quite different images. The standardization of medical equipment configurations in space missions would aid decrease the equipment induced domain shifts. 

Another domain shift source comes from the training patient characteristics. Different populations exhibit different medical characteristics. Hereditary traits as well as phenotype derived attributes place  an individual in different medical risk groups and force pathologies to manifest with different probabilities. These differences, are often picked up by ML algorithms as unwanted inductive biases, skewing the learned conclusions. In the context of space missions, astronauts have diverse backgrounds both genetically and in terms of phenotype, as such models trained on a general population not representative of the characteristics of the crew can provide skewed estimations of vital for diagnosis biomarkers. It is imperative, then, that the deployed models are calibrated to the genetic background and phenotype of the astronauts.

\noindent\textbf{\textit{Anomalous phenomena:}} As stated in the above motivation, the effects on human well-being stemming from prolonged exposure to cosmic radiation and other space related phenomena are not fully understood. It is unknown, hence how the human body might react to adverse conditions. ML applications cannot, then, be expected to cover the full range of scenarios, on the contrary they should be expected to fail when presented with data that constitute an anomalous effect. In order to combat potential failure cases of the medical systems on-board a spacecraft, we suggest accompanying any ML algorithm with an anomaly detection mechanism (perhaps probabilistic or out-of-distribution mechanics), that flags non standard bio markers~\cite{winkens2020contrastive}, and an active learning~\cite{settles_2012} feedback pipeline such that new, medical phenomena are incorporated in the evolution of the medical algorithms on-board.

\noindent\textbf{Computational Considerations} High communication latency creates the demand for on-board computation. However, cosmic radiation and weight offers strict constraints on the available on-board compute resources. Cosmic radiation is able to produce errors in modern scale computing units, while weight restrictions exist in all parts of a mission, from lift off to grounding. ML applications should, then, be able to function without heavy computational needs, a non trivial feat especially on medical imaging algorithms. This, reinforces our case towards the use of \gls{pocus} as they have been shown to be computationally and physically lightweight. Recently~\cite{9325840} has shown that NVIDIA's Jetson Xavier modules are able to withstand a significant level of proton based radiation, making them optimal candidates for on-board inference and fine-tuning infrastructure. NASA has also awarded an SBIR contract to Numem~\cite{HPSCDNN} to develop a radiation hardened DNN co-processor for a wide variety of ML applications, from low power machine vision to healthcare.

\section{Conclusion }

ML supported medical diagnostic systems on spacecraft are necessary for long-term space mission to overcome limitations in ground-to-spacecraft communication and lack of qualified medical crew. In this extended abstract, we outlined the importance of incorporating ML enabled medical applications on spacecrafts and considered challenges that need to be overcome in terms of anomaly detection and domain adaptation. Finally, we discussed ideas on how to augment existing \gls{pocus} algorithms such that they constitute a complete diagnostic system. We gave the example of a hierarchical RL method that also includes causal inference checks and balances such that the health and safety of crew members is guaranteed. We hope that with this paper we highlighted the final frontier for ML applications: space capable ML medical systems.

\section{Acknowledgments}
    We would like to thank NASA, the Canadian Space Agency (CSA), European Space Agency (ESA), NVIDIA, the SETI Institute and Frontier Development Lab (FDL) for their continuous support and for making this collaboration possible.




\bibliographystyle{named}
\bibliography{ijcai21.bib}

\end{document}